\documentclass{article}

\usepackage{microtype}
\usepackage{graphicx}
\usepackage{subcaption}
\usepackage{booktabs}
\usepackage{float}
\usepackage[table]{xcolor}
\usepackage{multirow}
\usepackage{makecell}

\usepackage{hyperref}

\usepackage[preprint]{icml2026}

\usepackage{amsmath}
\usepackage{amssymb}
\usepackage{mathtools}
\usepackage{amsthm}

\usepackage[capitalize,noabbrev]{cleveref}

\theoremstyle{plain}

\theoremstyle{definition}

\theoremstyle{remark}

\usepackage[textsize=tiny]{todonotes}

\icmltitlerunning{Submission and Formatting Instructions for ICML 2026}

\begin{document}

\twocolumn[
  \icmltitle{Entropy Reveals Block Importance in Masked Self-Supervised Vision Transformers}


  \icmlsetsymbol{equal}{*}

  \begin{icmlauthorlist}
    \icmlauthor{Peihao Xiang}{yyy}
    \icmlauthor{Kaida Wu}{yyy}
    \icmlauthor{Ou Bai}{yyy}
  \end{icmlauthorlist}

  \icmlaffiliation{yyy}{Department of Electrical and Computer Engineering, Florida International University, Miami, Florida, USA}

  \icmlcorrespondingauthor{Ou Bai}{obai@fiu.edu}

  \icmlkeywords{Self-Supervised Learning, Model Compression, Data-Free Pruning, Vision Transformers, Information-Theoretic Analysis, Efficient Transfer Learning}

  \vskip 0.3in
]



\printAffiliationsAndNotice{}  

\begin{abstract}

Masked self-supervised vision transformers have become a dominant pretraining paradigm, yet their substantial model size poses significant challenges for resource-constrained deployment and efficient transfer learning. A fundamental question remains: are all transformer blocks equally important for downstream performance? In this paper, we show that block importance in masked self-supervised vision transformers can be accurately estimated without access to any data. Our key finding is that the information entropy of pretrained block weights strongly correlates with oracle sensitivity obtained via iterative block removal and finetuning. This observation enables Gardener, a data-free, one-shot, block-level pruning principle that identifies redundant blocks through simple information-theoretic measurements. We evaluate Gardener on VideoMAE-B across multiple pruning ratios and downstream video recognition benchmarks. Despite its negligible computational overhead, Gardener consistently matches or outperforms existing data-free pruning baselines and closely approaches sensitivity-based pruning. Remarkably, even after pruning up to 91.7\% of blocks, the pruned model retains competitive transfer performance. Our results reveal substantial block-level redundancy in masked self-supervised vision transformers and demonstrate that information-theoretic analysis offers a principled and efficient pathway for model compression and resource-efficient transfer learning.
\end{abstract}

\section{Introduction}

Masked self-supervised learning has emerged as a dominant paradigm for pretraining large-scale vision transformers \cite{c1}, enabling strong transfer performance across a wide range of downstream tasks. Models such as MAE \cite{c2} and VideoMAE \cite{c3} learn rich visual representations from unlabeled data and have become foundational components in modern vision systems. However, the substantial size of these pretrained models poses significant challenges for deployment in resource-constrained environments, including edge devices, mobile platforms, and memory-limited local systems.

A natural approach to address this challenge is model compression \cite{c4}, among which pruning is particularly attractive due to its potential to reduce model size and computational cost without modifying training objectives or inference pipelines. Nevertheless, existing pruning methods \cite{c5} face fundamental limitations when applied to masked self-supervised vision transformers. Sensitivity-based pruning methods \cite{c6}, while effective, require iterative block removal and full finetuning, incurring prohibitive computational overhead. Data-driven pruning approaches \cite{c7} rely on downstream data or activation statistics, introducing dataset bias and limiting applicability in scenarios where data is unavailable or restricted. Moreover, unstructured pruning \cite{c8} often leads to irregular sparsity patterns that are difficult to deploy efficiently on modern hardware.

\begin{figure}[t]
    \centering
    \includegraphics[width=\linewidth]{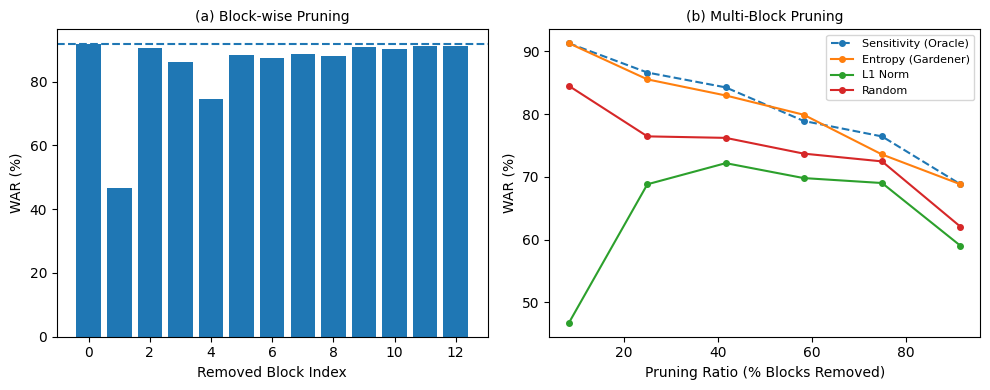}
    \caption{Pruning behavior of masked self-supervised vision transformers on VideoMAE-B finetuned on UCF101. (a) Block-wise pruning results, where index 0 denotes the original model and indices 1–12 correspond to removing individual transformer blocks. The dashed line indicates the unpruned baseline. The results reveal substantial block-level heterogeneity. (b) Multi-block pruning performance under increasing pruning ratios. Entropy-based pruning (Gardener) consistently tracks sensitivity-based (oracle) pruning and outperforms other data-free criteria across a wide range of pruning ratios.}
    \label{fig:pruned_results}
\end{figure}

These limitations raise a fundamental question: are all transformer \cite{c9} blocks equally important for transfer learning in masked self-supervised vision transformers, and can block importance be estimated efficiently without access to data? To illustrate this phenomenon, Figure 1 provides a block-level pruning analysis on VideoMAE-B \cite{c3}. As shown in Figure 1(a), removing different transformer \cite{c9} blocks leads to drastically different downstream performance, revealing substantial block-level heterogeneity. Figure 1(b) further shows that entropy-based block ranking closely tracks sensitivity-based (oracle) pruning under increasing pruning ratios, motivating a principled and data-free approach to block importance estimation.

In this work, we provide a negative answer to the first question and a positive answer to the second. Through extensive block-wise analysis, we show that transformer \cite{c9} blocks contribute unequally to downstream performance and that removing different blocks leads to drastically different accuracy degradation. While sensitivity-based pruning \cite{c6} can reveal this heterogeneity, its high computational cost makes it impractical as a general solution. Our key insight is that block importance is encoded in the pretrained parameters themselves. Specifically, we find that the weight-number entropy of a transformer \cite{c9} block—an information-theoretic measure of how parameter counts are distributed across value intervals—strongly correlates with oracle sensitivity obtained via iterative block removal and finetuning. This observation enables a principled, data-free surrogate for block importance estimation that operates solely on pretrained weights.

Based on this insight, we propose Gardener, a data-free, one-shot, block-level pruning method for masked self-supervised vision transformers. Gardener ranks blocks using weight-number entropy and removes redundant blocks in a single pass, without access to data, gradients, or iterative finetuning. Despite its negligible computational overhead, Gardener consistently matches or closely approaches sensitivity-based pruning and outperforms existing data-free baselines across both block-wise and multi-block pruning settings. Beyond practical performance, our analysis reveals systematic structural heterogeneity across transformer blocks in masked self-supervised models. Weight-number entropy captures intrinsic representational properties learned during pretraining, providing insight into why certain blocks are critical for transfer learning while others are redundant. We summarize our contributions as follows:
\begin{itemize}
  \item We formulate the problem of data-free block importance estimation for masked self-supervised vision transformers under resource-constrained deployment settings.
  \item We introduce weight-number entropy, an information-theoretic criterion computed solely from pretrained parameters and demonstrate its strong correlation with oracle sensitivity.
  \item We propose Gardener, a data-free, one-shot, block-level pruning algorithm with linear-time complexity and negligible memory overhead.
  \item We conduct extensive experiments on VideoMAE-B \cite{c3}, showing that Gardener consistently outperforms existing data-free pruning methods and achieves near-oracle performance under both block-wise and multi-block pruning.
  \item We provide empirical and conceptual analysis that sheds light on the internal organization and redundancy of masked self-supervised vision transformers.
\end{itemize}

\section{Related Work}
Model pruning \cite{c5} has been extensively studied to reduce the computational and memory footprint of deep neural networks. Early work primarily focused on unstructured pruning \cite{c8} at the weight level, which achieves high compression ratios but often leads to irregular sparsity patterns that are difficult to exploit efficiently on modern hardware. More recent efforts have shifted toward structured pruning \cite{c10}, where entire filters, layers, or blocks are removed to preserve architectural regularity and deployment efficiency.

\subsection{Structured Pruning}
Structured pruning \cite{c10} methods aim to remove coherent components of a model, such as channels, layers, or blocks, while maintaining performance. In convolutional networks, filter- and channel-level pruning has been widely explored. In contrast, Vision Transformers \cite{c1} introduce a natural modular structure composed of sequential transformer \cite{c9} blocks, making block-level pruning a particularly suitable strategy. Several works have investigated layer-wise or block-wise pruning \cite{c11,c12} for transformers, often relying on training-time signals or activation statistics to assess importance. However, these methods typically require access to data and repeated evaluation, limiting their applicability in data-restricted or resource-constrained settings.

\subsection{Data-Driven Pruning}
Sensitivity-based pruning evaluates the importance of model components by measuring the performance impact of their removal, often using gradients or second-order approximations \cite{c6}. While such methods provide strong performance guarantees and can be regarded as oracle criteria, they require iterative pruning and full retraining or finetuning, resulting in substantial computational overhead. Data-driven pruning approaches \cite{c7} estimate importance based on activation statistics, feature saliency, or downstream loss signals. Although effective in many settings, these approaches depend on representative data samples and may suffer from dataset bias or privacy constraints, particularly when downstream data is unavailable at pruning time.

\subsection{Data-Free Pruning}
Data-free pruning methods \cite{c13} seek to eliminate the reliance on training or validation data by operating directly on pretrained model parameters. Existing approaches often use magnitude-based criteria \cite{c14}, such as weight norms, variance, or higher-order moments, to estimate importance. While computationally efficient, these statistics provide limited insight into the structural role of model components and often exhibit unstable behavior across architectures or pruning ratios \cite{c15,c16}. As a result, their ability to approximate oracle sensitivity remains limited.

\subsection{Information-Theoretic Analysis}
Information-theoretic measures, including entropy and mutual information \cite{c17}, have been used to analyze representation learning and model behavior in neural networks. Prior work has explored entropy-based criteria for feature selection, compression, and interpretability \cite{c18,c19}, often in conjunction with data-dependent signals such as activations or learned representations \cite{c20}. In contrast, the use of information-theoretic analysis directly on pretrained parameters for structured pruning remains underexplored, particularly in the context of masked self-supervised vision transformers.

Our work bridges these lines of research by introducing a data-free, block-level pruning method grounded in information-theoretic analysis of pretrained parameters. Unlike sensitivity-based approaches \cite{c6}, our method avoids iterative evaluation and finetuning. Unlike existing data-free criteria \cite{c5,c13}, weight-number entropy captures the structural distribution of parameters within transformer blocks, enabling a closer approximation to oracle block importance. To the best of our knowledge, this is the first work to demonstrate that information-theoretic analysis of pretrained weights alone can reliably guide block-level pruning in masked self-supervised vision transformers.

\section{Method}

\subsection{Problem Formulation}
We consider a masked self-supervised vision transformer pretrained on unlabeled data, consisting of a sequence of $L$ transformer blocks, denoted as $\{B_1,...,B_L\}$. Given the pretrained model parameters, our goal is to identify a subset of blocks to be removed under a target resource budget, such that the resulting pruned model preserves maximal transfer performance after downstream finetuning.

Crucially, we focus on a data-free pruning setting, where no training or validation data from downstream tasks is accessible during the pruning decision process. This setting reflects practical deployment scenarios in which pretrained models must be adapted prior to task specification or data availability and avoids dataset-induced bias commonly introduced by data-driven pruning criteria.

Formally, let $S \subseteq \{1,...,L\}$ denote the set of blocks to be removed. The objective is to select $S$ such that the pruned model achieves optimal downstream performance subject to a resource constraint, while the selection of $S$ depends only on the pretrained model parameters.

The central challenge lies in estimating the importance of each transformer block using solely pretrained weights, without iterative finetuning, gradient-based sensitivity analysis, or data-dependent evaluation. Addressing this challenge requires a computationally efficient surrogate for block importance that can operate entirely in a data-free and one-shot manner. We seek a block importance scoring function $f(B_l)$ that ranks blocks according to their contribution to downstream transfer performance, under the constraint that $f$ operates solely on pretrained parameters. In this work, we treat sensitivity-based pruning as an oracle criterion and seek a low-cost surrogate that approximates such oracle behavior under zero-data constraints.

\subsection{Block-Level Weight Entropy}
To estimate block importance under the data-free constraint, we introduce block-level weight-number entropy, an information-theoretic measure computed solely from pretrained model parameters. Unlike conventional entropy definitions that operate on weight magnitudes or activations \cite{c17}, our formulation characterizes the distribution of parameter counts within a transformer block.

Consider a transformer block $B_l$ with parameter set $W_l$, containing $N_l=|W_l|$ scalar weights. We partition the range of weight values into a finite set of disjoint intervals $K$. Let $n_l(i)$ denote the number of weights in block $B_l$ whose values fall into interval $i$. We define empirical probability mass function as
\begin{equation}
\textbf{$p_l(i)$} = \frac{n_l(i)}{N_l}
\label{eq:prob}
\end{equation}

where the denominator corresponds to the total number of parameters in the block, rather than the sum of weight magnitudes.
The weight-number entropy of block $B_l$ is then defined as
\begin{equation}
\textbf{$H(B_l)$} = - \sum_i p_l(i)logp_l(i)
\label{eq:entropy}
\end{equation}

This formulation measures how uniformly parameter values are distributed across intervals, capturing the structural diversity of weights within each block while remaining invariant to weight scaling. By relying on parameter counts rather than magnitudes, weight-number entropy avoids biases introduced by a small number of large-magnitude weights and provides a stable, architecture-agnostic block importance score.

In the data-free pruning setting, entropy values are computed independently for each block and used to rank block importance. Blocks with lower weight-number entropy are considered less critical and are preferentially removed under a given pruning budget. Weight-number entropy can be computed in a single pass over pretrained parameters, incurs negligible computational overhead, and requires neither forward or backward propagation nor access to input data. We observe that block rankings induced by weight-number entropy are robust to reasonable choices of binning strategies and interval resolutions. The mathematical formulation for important critical ranking is as follows:
\begin{equation}
\textbf{$Rank(B_l)$} = \text{Sort}\{H(B_1),...,H(B_l)\}
\label{eq:rank}
\end{equation}

\begin{figure*}[t]
	\centering
	\includegraphics[scale=0.18]{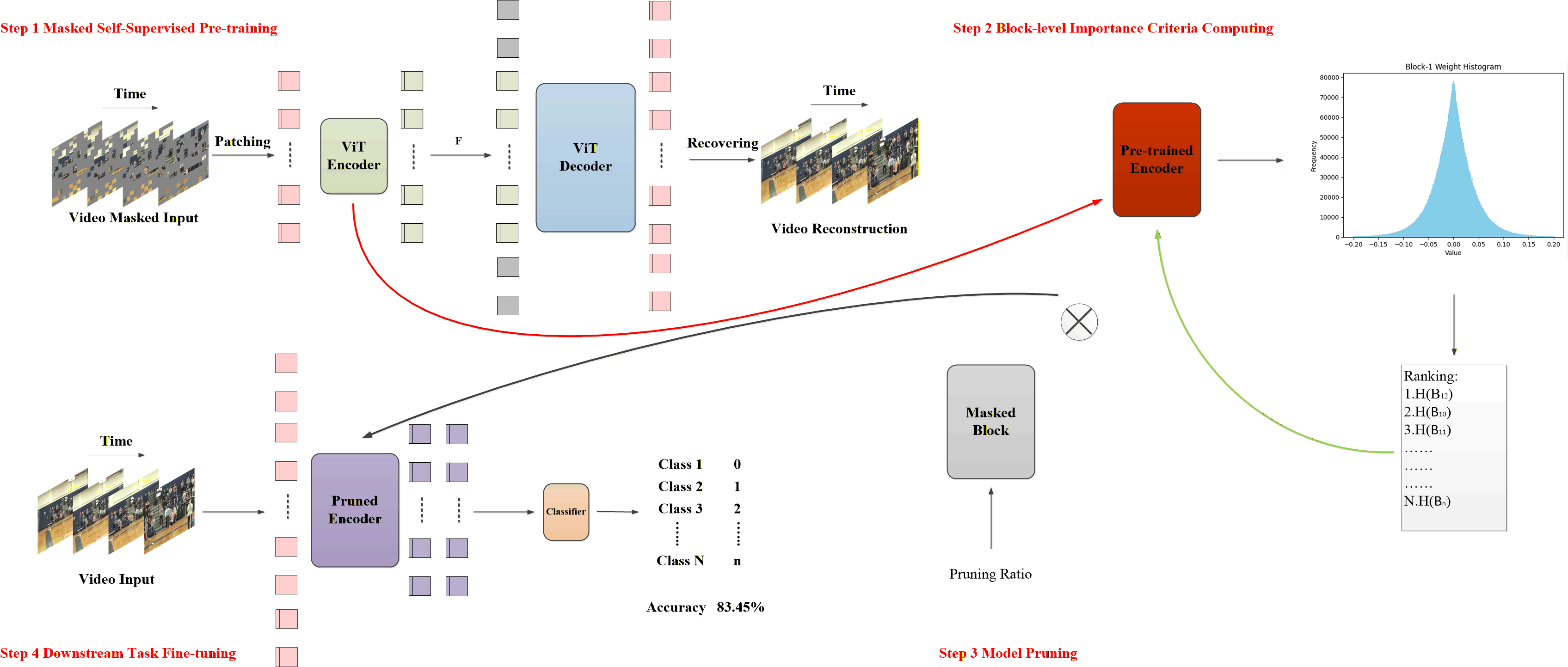}
	\caption{Architectural diagram of the information entropy-based Block-level Gardener pruning algorithm. Step 1: Visual self-supervised learning; Step 2: Calculating pruning criteria at the Block-level; Step 3: Pruning process of the VideoMAE Encoder Pretrained Model; Step 4: Fine-tuning the Pruned VideoMAE Encoder.}
	\label{fig:structure}
\end{figure*}

\subsection{Gardener Pruning Algorithm}
Building upon the block-level weight-number entropy defined in Section 3.2, we propose Gardener as shown in Algorithm 1, a data-free and one-shot pruning algorithm for masked self-supervised vision transformers. Gardener identifies and removes redundant transformer blocks solely based on pretrained model parameters, without requiring any data, gradient computation, or iterative finetuning.

Given a pretrained model consisting of L transformer blocks $\{B_1,...,B_L\}$, Gardener first computes the weight-number entropy $H(B_l)$ for each block independently. These entropy values serve as block importance scores. Blocks are then ranked in ascending order of entropy, where lower-entropy blocks are considered less critical to downstream transfer performance.

Let S denote the set of blocks to be removed under a target pruning budget. In contrast to iterative pruning schemes \cite{c6}, Gardener performs one-shot selection by directly choosing the lowest-ranked blocks according to entropy scores. This design avoids repeated model modification and evaluation, enabling efficient pruning even under strict computational constraints.

Gardener operates at the block level, removing entire transformer blocks while preserving the remaining model structure. This structured pruning strategy maintains architectural integrity and is naturally compatible with modern hardware accelerators and deployment frameworks as shown in Figure 2. After block removal, the pruned model is fine-tuned on downstream tasks using standard training procedures. Importantly, Gardener does not alter the downstream optimization process and can be seamlessly integrated with existing transfer learning pipelines \cite{c21}.

Gardener requires a single pass over pretrained parameters to compute block entropy scores, resulting in a time complexity $O(N_l)$ linear in the number of model parameters. No forward or backward propagation is required during pruning, making Gardener orders of magnitude more efficient than sensitivity-based pruning methods \cite{c6}. In contrast to sensitivity-based pruning methods \cite{c6} that require iterative finetuning and gradient computation, Gardener provides an efficient surrogate for block importance estimation under strict resource constraints.
Overall, Gardener provides a simple yet effective mechanism for data-free block pruning, enabling rapid model adaptation and resource-efficient deployment of masked self-supervised vision transformers. We treat sensitivity-based block pruning as an oracle criterion and view Gardener as a low-cost surrogate that approximates such oracle behavior under data-free constraints.

\begin{algorithm}[t]
\caption{Gardener Pruning Algorithm}
\label{alg:gardener}
\begin{algorithmic}[1]
\STATE \textbf{Input}: \text{Pretrained transformer blocks} $\{B_1, \ldots, B_L\}$ \\
                        \text{pruning ratio} $r \in (0,1)$ \\
\STATE \textbf{Parameter}: $l$ \\
\STATE \textbf{Output}: \text{Pruned transformer model} $M_{\text{pruned}}$
\FOR{$l = 1$ to $L$}
    \STATE \text{Extract all scalar weights} $N_l=|W_l|$ from block $B_l$
    \STATE \text{Partition weight values into} $K$ \text{disjoint intervals}
    \STATE \text{Count number of weights} $n_l(i)$ \text{in each interval} $i$
    \STATE \text{Compute empirical probability} $p_l(i) = n_l(i) / N_l$
    \STATE \text{Compute block entropy:} \\ $H(B_l) = - \sum_i p_l(i)\log p_l(i)$
\ENDFOR

\STATE \text{Rank blocks in ascending order of} $H(B_l)$
\STATE \text{Select the lowest-ranked} $\lfloor rL \rfloor$ blocks as pruning set $S$
\STATE \text{Remove blocks in} $S$ from the pretrained model
\STATE \textbf{Return} $M_{\text{pruned}}$
\end{algorithmic}
\end{algorithm}

\section{Experiments}

\subsection{Experimental Setup}
We conduct all experiments using a VideoMAE-Base \cite{c3} model pretrained on the Kinetics-400 dataset \cite{c22}. The pretrained model follows a Vision Transformer (ViT) \cite{c1} backbone and consists of 12 transformer \cite{c9} blocks, with a total of 86.6M parameters (approximately 330 MB). The input to the model is a sequence of 16 RGB video frames with spatial resolution $224*224$.

We evaluate transfer performance on the human action recognition task using the UCF101 dataset \cite{c23}. UCF101 contains 101 action categories and 10,100 videos, split into 7,070 training videos, 2,020 validation videos, and 1,010 test videos \cite{c23}. Following standard practice, no additional data augmentation is applied during finetuning, and Label Ranker \cite{c24} is used to eliminate the bias introduced by label position encoding.

After pruning, the resulting models are finetuned on UCF101 \cite{c23} using the same training configuration across all methods. We adopt the AdamW \cite{c25} optimizer with a weight decay of 0.05. The initial learning rate is set to $1 \times 10^{-3}$ and scheduled using a Warm-up followed by Cosine Decay \cite{c26}. Models are trained for 100 epochs, with the first 5 epochs used for learning rate Warm-up. The batch size is fixed to 64 for all experiments.

We report performance using Weighted Average Recall (WAR), which accounts for class imbalance and is commonly used in action recognition benchmarks. All reported results are evaluated on the test split of UCF101 \cite{c23}. All experiments are conducted on a V2-8 TPU with high-memory configuration. Importantly, pruning decisions are made entirely without access to downstream data or training signals, and computational resources are only used during the standard finetuning stage.

\subsection{Oracle Analysis}
We evaluate whether weight-number entropy provides a reliable surrogate for block importance by comparing it against sensitivity-based pruning \cite{c6}, which we treat as an oracle criterion. Sensitivity-based analysis measures the impact of removing transformer \cite{c9} blocks on downstream performance but requires iterative block removal and full finetuning, resulting in high computational cost.

We first conduct block-wise pruning experiments by removing a single transformer \cite{c9} block at a time from a pretrained VideoMAE-B \cite{c3} model and finetuning the resulting model on UCF101 \cite{c23}. The performance degradation induced by removing each block serves as a direct estimate of block sensitivity. As shown in Table 1, removing different blocks leads to substantially different downstream accuracies, confirming that transformer \cite{c9} blocks are not equally important for transfer learning.

We then rank all blocks using weight-number entropy computed from pretrained parameters and compare this ranking with the oracle sensitivity ranking derived from block-wise pruning as shown in Table 2. We observe a strong alignment between the two rankings: blocks with higher weight-number entropy consistently correspond to larger performance drops when removed, whereas blocks with lower entropy exhibit smaller impact on downstream accuracy. This result indicates that weight-number entropy captures intrinsic properties of block importance that are otherwise revealed only through expensive oracle evaluation.

To further validate the correlation between entropy-based importance estimation and oracle sensitivity, we evaluate multi-block pruning under identical pruning ratios using different data-free criteria. As summarized in Table 3, entropy-based pruning consistently outperforms alternative data-free baselines, including random selection \cite{c14}, mean-based, variance-based, norm-based, and kurtosis-based pruning \cite{c5} etc. Across pruning ratios, entropy-based pruning achieves downstream performance closest to sensitivity-based pruning \cite{c6}, despite operating in a one-shot and data-free manner.

Taken together, the block-wise and multi-block results provide strong empirical evidence that weight-number entropy serves as a reliable surrogate for sensitivity-based block importance. Importantly, this alignment holds across both fine-grained (single-block) and coarse-grained (multi-block) pruning settings, validating the effectiveness of entropy-based importance estimation beyond isolated cases. These findings justify the use of entropy as the core criterion in Gardener and explain its ability to achieve near-oracle pruning performance with negligible computational overhead.

\begin{table}[t]
\begin{center}
\centering
\caption{Block-wise pruning performance of VideoMAE-B finetuned on UCF101. RB: Remove Block. See A.1 for details.}
\label{tab:blockwise}
\begin{tabular}{lcccc}
\toprule
Model State & Params (M) & GFLOPs & WAR (\%) \\
\midrule
Original        & 86.6 & 180 & 91.78 \\
RB-1  & 79.5 & 166 & 46.73 \\
RB-2  & 79.5 & 166 & 90.58 \\
RB-3  & 79.5 & 166 & 86.14 \\
RB-4  & 79.5 & 166 & 74.65 \\
RB-5  & 79.5 & 166 & 88.32 \\
RB-6  & 79.5 & 166 & 87.43 \\
RB-7  & 79.5 & 166 & 88.51 \\
RB-8  & 79.5 & 166 & 88.02 \\
RB-9  & 79.5 & 166 & 90.79 \\
RB-10 & 79.5 & 166 & 90.30 \\
RB-11 & 79.5 & 166 & 91.09 \\
RB-12 & 79.5 & 166 & \textbf{91.29} \\
\bottomrule
\end{tabular}
\end{center}
\end{table}

\subsection{Results}
We now evaluate the effectiveness of Gardener under both block-wise and multi-block pruning settings and analyze how different data-free criteria align with oracle sensitivity. All experiments are conducted on VideoMAE-B \cite{c3} pretrained on Kinetics-400 \cite{c22} and finetuned on UCF101 \cite{c23}.

We first remove a single transformer \cite{c9} block at a time and finetune the resulting model. The downstream accuracies are reported in Table 1. Removing different blocks leads to substantial performance variation, indicating that transformer \cite{c9} blocks contribute unequally to transfer performance. Removing certain blocks results in dramatic accuracy degradation, while removing others has a relatively minor impact. These results define block-wise sensitivity, which we treat as an oracle measure of block importance.

To better understand how different data-free criteria approximate oracle sensitivity, we summarize block-wise statistics and ranking results in Table 2. The table reports block-wise accuracy, entropy-based scores, mutual information scores, efficiency-related metrics, and their corresponding rankings. We observe that entropy-based ranking exhibits the strongest alignment with sensitivity ranking, consistently assigning higher importance to blocks whose removal leads to larger performance drops. In contrast, magnitude-based and efficiency-based criteria show weaker or less stable correspondence with oracle sensitivity.

We further evaluate Gardener under multi-block pruning by removing multiple blocks simultaneously under identical pruning ratios. The results are summarized in Table 3. Across all pruning settings, Gardener consistently outperforms alternative data-free pruning baselines and remains closest to sensitivity-based pruning \cite{c6}. These findings demonstrate that the superiority of entropy-based importance estimation extends beyond single-block analysis and generalizes to more aggressive multi-block pruning scenarios.

Overall, the block-wise, statistical, and multi-block results jointly confirm that weight-number entropy provides a robust and scalable surrogate for oracle sensitivity, enabling effective one-shot pruning without access to data or gradients.

\begin{table}[t]
\begin{center}
\centering
\caption{Comparison between entropy-based vs. sensitivity-based oracle ranking. See A.2 for details.}
\label{tab:ranking}
\begin{tabular}{lcccc}
\toprule
Block & WAR (\%) & Entropy Rank & Sensitivity Rank \\
\midrule
1  & 46.73 & 1  & 1  \\
4  & 74.65 & 3  & 2  \\
3  & 86.14 & 5  & 3  \\
6  & 87.43 & 4  & 4  \\
5  & 88.32 & 6  & 6  \\
7  & 88.51 & 8  & 7  \\
8  & 88.02 & 7  & 5  \\
2  & 90.58 & 2  & 9  \\
9  & 90.79 & 9  & 10 \\
10 & 90.30 & 11 & 8  \\
11 & 91.09 & 10 & 11 \\
12 & 91.29 & 12 & 12 \\
\bottomrule
\end{tabular}
\end{center}
\end{table}

\begin{table}[t]
\begin{center}
\centering
\caption{Multi-block pruning performance under different pruning ratios on UCF101.PR: Pruning Ratio. See A.3 for details.}
\label{tab:multiblock}
\begin{tabular}{lcccc}
\toprule
\multirow{2}{*}{PR (\%)} & \multicolumn{4}{c}{WAR (\%)} \\
\cmidrule(lr){2-5}
 & Sensitivity & Entropy & L1 Norm & Random \\
\midrule
8.3  & \textbf{91.29} & \textbf{91.29} & 46.73 & 84.49 \\
25.0 & \textbf{86.63} & 85.54 & 68.81 & 76.45 \\
41.7 & \textbf{84.26} & 82.97 & 72.18 & 76.21 \\
58.3 & 78.91 & \textbf{79.90} & 69.80 & 73.70 \\
75.0 & \textbf{76.44} & 73.56 & 69.01 & 72.46 \\
91.7 & \textbf{68.81} & \textbf{68.81} & 59.01 & 62.05 \\
\bottomrule
\end{tabular}
\end{center}
\end{table}

\section{Discussion}
This section discusses why weight-number entropy is an effective block importance indicator, what it reveals about masked self-supervised vision transformers, and the limitations of the proposed approach. Our results show that block-level weight-number entropy strongly correlates with sensitivity-based block importance, despite being computed solely from pretrained parameters. This effectiveness stems from the fact that weight-number entropy captures how parameters are structurally allocated within a transformer \cite{c9} block. Unlike magnitude-based statistics, which focus on the scale of weights, entropy measures the distribution of parameter counts across value intervals, reflecting the diversity of representational pathways encoded by a block.

Empirically, we observe that blocks with higher weight-number entropy consistently induce larger performance degradation when removed, indicating higher importance for downstream transfer. Notably, this relationship holds even when visual inspection of weight distributions suggests sharp concentration as shown in Figure 3. This highlights a key distinction between apparent value concentration and entropy: a block may exhibit sharp peaks while still maintaining a broadly distributed allocation of parameters across intervals, resulting in high entropy and strong representational capacity. Sensitivity-based pruning \cite{c6} evaluates block importance by measuring the downstream performance drop caused by block removal and can be regarded as an oracle criterion. However, it requires iterative block removal and full finetuning, making it computationally prohibitive. Our experiments demonstrate that weight-number entropy provides a low-cost surrogate that closely approximates sensitivity-based rankings at the block level. This alignment suggests that entropy captures intrinsic properties of pretrained representations that govern their contribution to transfer learning, even before any downstream data is observed.

The observed entropy patterns offer insights into the internal organization of masked self-supervised vision transformers. Early and intermediate blocks tend to exhibit higher weight-number entropy, reflecting their role in capturing diverse low-level and mid-level visual patterns under reconstruction objectives. Deeper blocks show progressively lower entropy, indicating increased specialization toward semantic abstraction. This structural heterogeneity explains why blocks are not equally important for transfer learning and motivates block-level pruning as a natural and effective compression strategy for ViT-based \cite{c1} masked autoencoders.

Despite its effectiveness, Gardener has several limitations. First, weight-number entropy is a coarse-grained, static measure that does not account for input-dependent behavior or task-specific activation patterns. While this design is intended to enable data-free pruning, it may limit adaptability to highly specialized downstream tasks. Second, the current study focuses on block-level pruning in ViT-based \cite{c1} masked self-supervised models; extending entropy-based criteria to finer-grained structures such as attention heads or MLP \cite{c11,c12} sublayers remains an open question. Finally, although entropy correlates well with sensitivity in our experiments, it is not guaranteed to perfectly approximate oracle importance in all architectures or training regimes. These limitations suggest several promising directions for future work, including combining entropy-based criteria with lightweight data-aware signals, extending the approach to other self-supervised modalities such as audio or multimodal transformers, and exploring entropy-aware model design that explicitly encourages redundancy where compression is desirable. We believe that information-theoretic analysis of pretrained parameters offers a principled pathway toward more efficient and interpretable foundation models.

\begin{figure}[t]
    \centering
    \includegraphics[width=\linewidth]{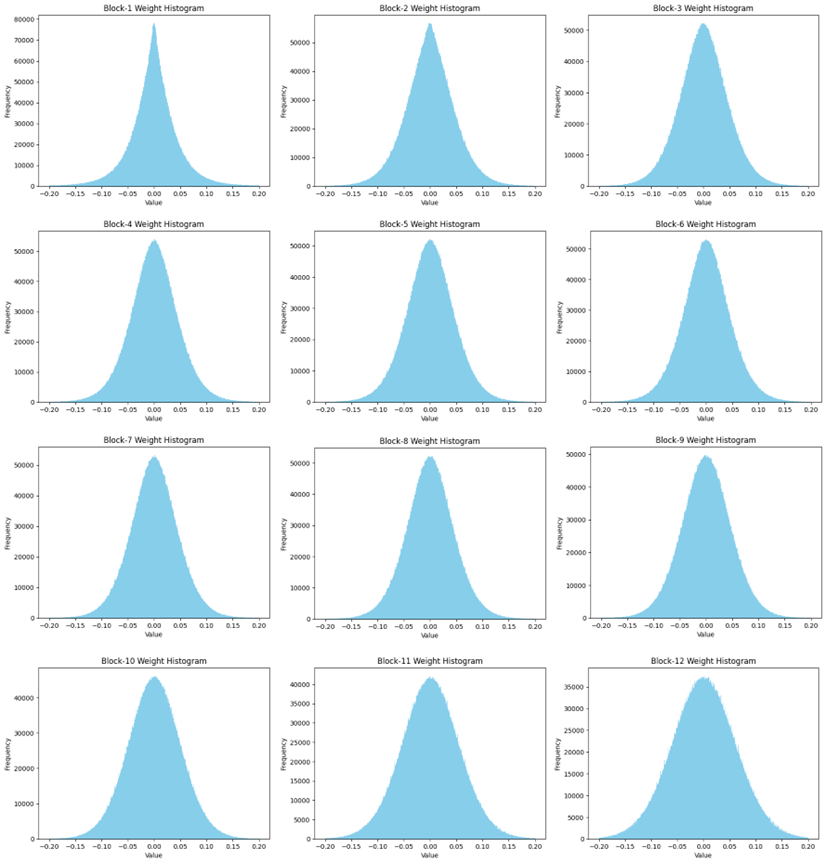}
    \caption{Weight distributions of transformer blocks at different depths in a pretrained VideoMAE encoder. Early blocks exhibit sharply peaked distributions, while deeper blocks show increasingly flatter and more dispersed parameter distributions.}
    \label{fig:block_hist}
\end{figure}

\section{Conclusion}
In this work, we studied the problem of block importance estimation in masked self-supervised vision transformers under strict data-free and resource-constrained settings. We showed that transformer \cite{c9} blocks contribute unequally to downstream transfer performance and that identifying redundant blocks is critical for efficient deployment.

To address this challenge, we introduced Gardener, a data-free, one-shot, block-level pruning method based on weight-number entropy. By analyzing the distribution of pretrained parameters, Gardener provides a computationally efficient surrogate for sensitivity-based pruning without requiring access to data, gradients, or iterative finetuning. Extensive experiments on VideoMAE-B \cite{c3} demonstrate that entropy-based block ranking strongly correlates with oracle sensitivity and consistently outperforms existing data-free pruning criteria across both block-wise and multi-block pruning settings.

Beyond empirical performance, our analysis reveals systematic structural heterogeneity across transformer blocks in masked self-supervised models. Weight-number entropy captures intrinsic representational properties encoded during pretraining, offering insight into why certain blocks are more critical for transfer learning than others.

Overall, this work highlights the potential of information-theoretic analysis of pretrained parameters as a principled approach to model compression. We hope that our findings will inspire further research on data-free and resource-efficient adaptation of large-scale self-supervised models for real-world deployment.

\section*{Impact Statement}
This paper presents work whose goal is to advance the field of Machine
Learning. There are many potential societal consequences of our work, none
which we feel must be specifically highlighted here.

\nocite{langley00}

\bibliography{example_paper}
\bibliographystyle{icml2026}

\newpage
\appendix
\onecolumn

\section{Additional Experimental Results}

\subsection{Full Block-wise Pruning Results}

\begin{table}[H]
\centering
\caption{Full block-wise pruning results of VideoMAE-B finetuned on UCF101, including detailed training, validation, and test metrics for each block removal.}
\label{tab:block_removal_appendix}
\resizebox{\textwidth}{!}{
\begin{tabular}{lccccccccccc}
\toprule
Model State & Size (MB) & \#Params (M) & Train Loss & Train Acc. (\%) & Val Loss & Val Acc. (\%) & Test Loss & Test Acc. (\%) & GFLOPs & WAR (\%) \\
\midrule
Original
& 330.26 & 86.6
& 0.2223 & 95.20
& 0.2808 & 94.91
& 0.3608 & 91.67
& 180 & 91.78 \\

Remove Block-1
& 303.22 & 79.5
& 1.9419 & 47.00
& 1.9590 & 47.73
& 1.9471 & 47.40
& 166 & 46.73 \\

Remove Block-2
& 303.22 & 79.5
& 0.0716 & 97.78
& 0.3205 & 91.33
& 0.3009 & 90.77
& 166 & 90.58 \\

Remove Block-3
& 303.22 & 79.5
& 0.1481 & 95.57
& 0.4028 & 88.91
& 0.4556 & 86.04
& 166 & 86.14 \\

Remove Block-4
& 303.22 & 79.5
& 0.7273 & 78.84
& 0.8952 & 75.50
& 0.8930 & 74.79
& 166 & 74.65 \\

Remove Block-5
& 303.22 & 79.5
& 0.1647 & 94.94
& 0.3836 & 88.86
& 0.3981 & 88.54
& 166 & 88.32 \\

Remove Block-6
& 303.22 & 79.5
& 0.2022 & 93.72
& 0.4436 & 87.70
& 0.4581 & 87.50
& 166 & 87.43 \\

Remove Block-7
& 303.22 & 79.5
& 0.0997 & 96.95
& 0.3459 & 90.02
& 0.4151 & 88.65
& 166 & 88.51 \\

Remove Block-8
& 303.22 & 79.5
& 0.1119 & 96.95
& 0.3991 & 89.06
& 0.4002 & 87.81
& 166 & 88.02 \\

Remove Block-9
& 303.22 & 79.5
& 0.0588 & 98.20
& 0.2989 & 91.89
& 0.3278 & 90.42
& 166 & 90.79 \\

Remove Block-10
& 303.22 & 79.5
& 0.0792 & 97.68
& 0.2927 & 92.49
& 0.3468 & 90.52
& 166 & 90.30 \\

Remove Block-11
& 303.22 & 79.5
& 0.0718 & 97.90
& 0.3131 & 90.78
& 0.3046 & 91.25
& 166 & 91.09 \\

\rowcolor{yellow!50}
Remove Block-12
& 303.22 & 79.5
& 0.0183 & \textbf{99.76}
& 0.3110 & \textbf{91.78}
& 0.3297 & \textbf{91.46}
& 166 & \textbf{91.29} \\

\bottomrule
\end{tabular}
}
\end{table}

Due to space constraints, we provide the full block-wise pruning results and detailed training statistics in Table~\ref{tab:block_removal_appendix}. These results include training, validation, and test metrics for each block removal setting.

\subsection{Full Block-wise Statistical Analysis}

Table~\ref{tab:block_statistics_part1} and Table~\ref{tab:block_statistics_part2} reports detailed block-wise statistics, including entropy, mutual information, magnitude-based metrics, and their corresponding rankings.

\begin{table}[H]
\centering
\caption{Statistical comparison of block-wise weights for VideoMAE-B (K400 training).
This table reports performance, information-theoretic metrics, and efficiency-related statistics.}
\label{tab:block_statistics_part1}
\resizebox{\textwidth}{!}{
\begin{tabular}{lccccccccccccccc}
\toprule
Model & Acc. (\%) & $\Delta$ (\%) & Rank
& Entropy$_N$ & Rank
& MI$_N$ & Rank
& Efficiency & Rank
& Entropy$_V$ & Rank
& MI$_V$ & Rank
& Sensitive & Rank \\
\midrule
Original
& \textbf{91.78} & -- & --
& -- & --
& -- & --
& 0.5150 & --
& -- & --
& -- & --
& -- & -- \\

Block-1
& 46.73 & 45.05 & 1
& 1 & 1
& 1 & \cellcolor{yellow!50}\textbf{1}
& 0.6054 & 1
& 0 & 12
& 0 & 12
& 12.8957 & 1 \\

Block-2
& 90.58 & 1.20 & 9
& 0.3420 & 2
& 0.4653 & 2
& 0.5971 & 2
& 0.7617 & 11
& 0.7095 & 11
& 0.3252 & 9 \\

Block-3
& 86.14 & 5.64 & 3
& 0.1845 & 5
& 0.3161 & 5
& 0.5951 & 5
& 0.8746 & 8
& 0.8114 & 8
& 1.5121 & 3 \\

Block-4
& 74.65 & 17.13 & 2
& 0.2297 & 3
& 0.3559 & 3
& 0.5956 & 3
& 0.8296 & 10
& 0.7731 & 10
& 4.6835 & 2 \\

Block-5
& 88.32 & 3.46 & 6
& 0.1715 & 6
& 0.3027 & 6
& 0.5949 & 6
& 0.8782 & 7
& 0.8151 & 7
& 0.9307 & 6 \\

Block-6
& 87.43 & 4.35 & 4
& 0.2038 & 4
& 0.3322 & 4
& 0.5953 & 4
& 0.8533 & 9
& 0.7942 & 9
& 1.1842 & 4 \\

Block-7
& 88.51 & 3.27 & 7
& 0.1443 & 8
& 0.2710 & 8
& 0.5946 & 8
& 0.9018 & 5
& 0.8412 & 5
& 0.9003 & 7 \\

Block-8
& 88.02 & 3.76 & 5
& 0.1499 & 7
& 0.2794 & 7
& 0.5946 & 7
& 0.8946 & 6
& 0.8321 & 6
& 1.0233 & 5 \\

Block-9
& 90.79 & 0.99 & 10
& 0.0627 & 9
& 0.1963 & 9
& 0.5935 & 9
& 0.9626 & 4
& 0.8896 & 4
& 0.2622 & 10 \\

Block-10
& 90.30 & 1.48 & 8
& 0.0134 & 11
& 0.1436 & 10
& 0.5929 & 11
& 1 & \cellcolor{yellow!50}\textbf{1}
& 0.9202 & 3
& 0.3676 & 8 \\

Block-11
& 91.09 & 0.69 & 11
& 0.0262 & 10
& 0.1176 & 11
& 0.5931 & 10
& 0.9911 & 3
& 0.9341 & 2
& 0.1554 & 11 \\

Block-12
& 91.29 & \textbf{0.49} & \cellcolor{yellow!50}\textbf{12}
& 0 & \cellcolor{yellow!50}\textbf{12}
& 0 & 12
& 0.5927 & \cellcolor{yellow!50}\textbf{12}
& 0.9991 & 2
& 1 & \cellcolor{yellow!50}\textbf{1}
& 0.0991 & \cellcolor{yellow!50}\textbf{12} \\

\bottomrule
\end{tabular}
}
\end{table}

\begin{table}[H]
\centering
\caption{Distribution statistics of block-wise weights, including norm-based
and higher-order statistical measures.}
\label{tab:block_statistics_part2}
\resizebox{\textwidth}{!}{
\begin{tabular}{lccccccccccccccc}
\toprule
Model & Cot. zero
& Max & Rank
& Mean & Rank
& L1 Norm & Rank
& L2 Norm & Rank
& Var & Rank
& Std & Rank
& Kurtosis & Rank \\
\midrule
Original
& 46
& 1.7090 & --
& 0.0387 & --
& -- & --
& -- & --
& 0.001074 & --
& 0.0328 & --
& 3.1024 & -- \\

Block-1
& 9
& 1.7090 & 1
& 0.0349 & \cellcolor{yellow!50}\textbf{12}
& 0 & \cellcolor{yellow!50}\textbf{12}
& 0.3549 & 3
& 0.001544 & 1
& 0.0393 & 1
& 16.2497 & \cellcolor{yellow!50}\textbf{1} \\

Block-2
& 6
& 1.1221 & 8
& 0.0369 & 7
& 0.1354 & 7
& 0.1058 & 5
& 0.000996 & 4
& 0.0316 & 4
& 1.9431 & 5 \\

Block-3
& 2
& 0.9580 & 12
& 0.0373 & 5
& 0.1628 & 5
& 0.0887 & 6
& 0.000940 & 7
& 0.0307 & 7
& 1.3087 & 9 \\

Block-4
& 5
& 1.2158 & 6
& 0.0366 & 10
& 0.1130 & 10
& 0.0572 & 9
& 0.000945 & 6
& 0.0307 & 6
& 2.1245 & 3 \\

Block-5
& 4
& 0.9691 & 11
& 0.0372 & 6
& 0.1544 & 6
& 0.0811 & 8
& 0.000937 & 8
& 0.0306 & 8
& 1.5999 & 8 \\

Block-6
& 1
& 1.3604 & 3
& 0.0367 & 9
& 0.1240 & 9
& 0.0566 & 10
& 0.000932 & 9
& 0.0305 & 9
& 2.0342 & 4 \\

Block-7
& 4
& 1.2480 & 5
& 0.0363 & 11
& 0.0956 & 11
& 0      & \cellcolor{yellow!50}\textbf{12}
& 0.000876 & \cellcolor{yellow!50}\textbf{12}
& 0.0296 & \cellcolor{yellow!50}\textbf{12}
& 1.8137 & 6 \\

Block-8
& 1
& 1.1133 & 9
& 0.0367 & 8
& 0.1246 & 8
& 0.0386 & 11
& 0.000904 & 10
& 0.0301 & 10
& 1.6516 & 7 \\

Block-9
& 4
& 1.2988 & 4
& 0.0378 & 4
& 0.1940 & 4
& 0.0846 & 7
& 0.000900 & 11
& 0.0300 & 11
& 0.9015 & 10 \\

Block-10
& 5
& 1.1523 & 7
& 0.0403 & 3
& 0.3666 & 3
& 0.2633 & 4
& 0.000990 & 5
& 0.0315 & 5
& 0.5767 & 11 \\

Block-11
& 4
& 1.0361 & 10
& 0.0444 & 2
& 0.6527 & 2
& 0.5957 & 2
& 0.001214 & 3
& 0.0348 & 3
& 0.5308 & 12 \\

Block-12
& 1
& 1.4404 & 2
& 0.0495 & 1
& 1 & 1
& 1 & 1
& 0.001518 & 2
& 0.0390 & 2
& 2.1493 & 2 \\

\bottomrule
\end{tabular}
}
\end{table}

\subsection{Full Multi-block Pruning Strategies}

\begin{table}[H]
\centering

\begin{minipage}[t]{0.48\textwidth}
\centering
\caption{Pruning strategies when removing 1 block (8.30\%).}
\label{tab:appendix_remove1}
\resizebox{\columnwidth}{!}{
\begin{tabular}{clcccc}
\toprule
Pruning Ratio
& \multicolumn{1}{c}{Method}
& Removed Block
& GFLOPs
& WAR (\%)
& $\Delta$ (\%) \\
\midrule
\multirow{10}{*}{8.30\%}

& Random & --
& 166 & 84.49 & 7.29 \\

& Mean & 1
& 166 & 46.73 & 45.05 \\

& Variance & 7
& 166 & 88.51 & 3.27 \\

& L1 Norm & 1
& 166 & 46.73 & 45.05 \\

& \cellcolor{yellow!50}Sensitivity
& \cellcolor{yellow!50}12
& \cellcolor{yellow!50}166
& \cellcolor{yellow!50}91.29
& \cellcolor{yellow!50}0.49 \\

& L2 Norm & 7
& 166 & 88.51 & 3.27 \\

& Standard Deviation & 7
& 166 & 88.51 & 3.27 \\

& Kurtosis & 1
& 166 & 46.73 & 45.05 \\

\cmidrule(lr){2-6}
& \cellcolor{yellow!50}Information Entropy
& \cellcolor{yellow!50}12
& \cellcolor{yellow!50}166
& \cellcolor{yellow!50}91.29
& \cellcolor{yellow!50}0.49 \\

& Mutual Information & 1
& 166 & 46.73 & 45.05 \\

\bottomrule
\end{tabular}
}
\end{minipage}
\hfill
\begin{minipage}[t]{0.48\textwidth}
\centering
\caption{Pruning strategies when removing 3 blocks (25\%).}
\label{tab:appendix_remove3}
\resizebox{\columnwidth}{!}{
\begin{tabular}{clcccc}
\toprule
Pruning Ratio
& \multicolumn{1}{c}{Method}
& Removed Block
& GFLOPs
& WAR (\%)
& $\Delta$ (\%) \\
\midrule
\multirow{10}{*}{25\%}

& Random & --
& 139 & 76.45 & 15.33 \\

& Mean & 1,4,7
& 139 & 68.81 & 22.97 \\

& Variance & 7,8,9
& 139 & 85.25 & 6.53 \\

& L1 Norm & 1,4,7
& 139 & 68.81 & 22.97 \\

& \cellcolor{yellow!50}Sensitivity
& \cellcolor{yellow!50}9,11,12
& \cellcolor{yellow!50}139
& \cellcolor{yellow!50}86.63
& \cellcolor{yellow!50}5.15 \\

& L2 Norm & 6,7,8
& 139 & 83.56 & 8.22 \\

& Standard Deviation & 7,8,9
& 139 & 85.25 & 6.53 \\

& Kurtosis & 1,4,12
& 139 & 64.36 & 27.42 \\

\cmidrule(lr){2-6}
& \cellcolor{yellow!50}Information Entropy
& \cellcolor{yellow!50}10,11,12
& \cellcolor{yellow!50}139
& \cellcolor{yellow!50}85.54
& \cellcolor{yellow!50}6.24 \\

& Mutual Information & 1,2,4
& 139 & 60.99 & 30.79 \\

\bottomrule
\end{tabular}
}
\end{minipage}

\end{table}

\begin{table}[H]
\centering

\begin{minipage}[t]{0.48\textwidth}
\centering
\caption{Pruning strategies when removing 5 blocks (41.70\%).}
\label{tab:appendix_remove5}
\resizebox{\columnwidth}{!}{
\begin{tabular}{clcccc}
\toprule
Pruning Ratio
& \multicolumn{1}{c}{Method}
& Removed Block
& GFLOPs
& WAR (\%)
& $\Delta$ (\%) \\
\midrule
\multirow{10}{*}{41.70\%}

& Random & --
& 112 & 76.21 & 15.57 \\

& Mean & 1,4,6,7,8
& 112 & 72.18 & 19.60 \\

& Variance & 5,6,7,8,9
& 112 & 79.70 & 12.08 \\

& L1 Norm & 1,4,6,7,8
& 112 & 72.18 & 19.60 \\

& \cellcolor{yellow!50}Sensitivity
& \cellcolor{yellow!50}2,9,10,11,12
& \cellcolor{yellow!50}112
& \cellcolor{yellow!50}84.26
& \cellcolor{yellow!50}7.52 \\

& L2 Norm & 4,5,6,7,8
& 112 & 81.58 & 10.20 \\

& Standard Deviation & 5,6,7,8,9
& 112 & 79.70 & 12.08 \\

& Kurtosis & 1,2,4,6,12
& 112 & 70.30 & 21.48 \\

\cmidrule(lr){2-6}
& \cellcolor{yellow!50}Information Entropy
& \cellcolor{yellow!50}7,9,10,11,12
& \cellcolor{yellow!50}112
& \cellcolor{yellow!50}82.97
& \cellcolor{yellow!50}8.81 \\

& Mutual Information & 1,2,3,4,6
& 112 & 62.48 & 29.30 \\

\bottomrule
\end{tabular}
}
\end{minipage}
\hfill
\begin{minipage}[t]{0.48\textwidth}
\centering
\caption{Pruning strategies when removing 7 blocks (58.30\%).}
\label{tab:appendix_remove7}
\resizebox{\columnwidth}{!}{
\begin{tabular}{clcccc}
\toprule
Pruning Ratio
& \multicolumn{1}{c}{Method}
& Removed Block
& GFLOPs
& WAR (\%)
& $\Delta$ (\%) \\
\midrule
\multirow{10}{*}{58.30\%}

& Random & --
& 84 & 73.70 & 18.08 \\

& Mean & 1,2,4,5,6,7,8
& 84 & 69.80 & 21.98 \\

& Variance & 3,4,5,6,7,8,9
& 84 & 78.02 & 13.76 \\

& L1 Norm & 1,2,4,5,6,7,8
& 84 & 69.80 & 21.98 \\

& \cellcolor{yellow!50}Sensitivity
& \cellcolor{yellow!50}2,5,7,9,10,11,12
& \cellcolor{yellow!50}84
& \cellcolor{yellow!50}78.91
& \cellcolor{yellow!50}12.87 \\

& L2 Norm & 3,4,5,6,7,8,9
& 84 & 78.02 & 13.76 \\

& Standard Deviation & 3,4,5,6,7,8,9
& 84 & 78.02 & 13.76 \\

& Kurtosis & 1,2,4,6,7,8,12
& 84 & 70.40 & 21.38 \\

\cmidrule(lr){2-6}
& \cellcolor{yellow!50}Information Entropy
& \cellcolor{yellow!50}5,7,8,9,10,11,12
& \cellcolor{yellow!50}84
& \cellcolor{yellow!50}79.90
& \cellcolor{yellow!50}11.88 \\

& Mutual Information & 1,2,3,4,5,6,8
& 84 & 65.15 & 26.63 \\

\bottomrule
\end{tabular}
}
\end{minipage}

\end{table}

\begin{table}[H]
\centering

\begin{minipage}[t]{0.48\textwidth}
\centering
\caption{Pruning strategies when removing 9 blocks (75\%).}
\label{tab:appendix_remove9}
\resizebox{\columnwidth}{!}{
\begin{tabular}{clcccc}
\toprule
Pruning Ratio
& \multicolumn{1}{c}{Method}
& Removed Block
& GFLOPs
& WAR (\%)
& $\Delta$ (\%) \\
\midrule
\multirow{10}{*}{75\%}

& Random & --
& 57 & 72.46 & 19.32 \\

& Mean & 1,2,3,4,5,6,7,8,9
& 57 & 69.01 & 22.77 \\

& Variance & 2,3,4,5,6,7,8,9,10
& 57 & 72.97 & 18.81 \\

& L1 Norm & 1,2,3,4,5,6,7,8,9
& 57 & 69.01 & 22.77 \\

& \cellcolor{yellow!50}Sensitivity
& \cellcolor{yellow!50}2,5,6,7,8,9,10,11,12
& \cellcolor{yellow!50}57
& \cellcolor{yellow!50}76.44
& \cellcolor{yellow!50}15.34 \\

& L2 Norm & 2,3,4,5,6,7,8,9,10
& 57 & 72.97 & 18.81 \\

& Standard Deviation & 2,3,4,5,6,7,8,9,10
& 57 & 72.97 & 18.81 \\

& Kurtosis & 1,2,3,4,5,6,7,8,12
& 57 & 70.30 & 21.48 \\

\cmidrule(lr){2-6}
& \cellcolor{yellow!50}Information Entropy
& \cellcolor{yellow!50}3,5,6,7,8,9,10,11,12
& \cellcolor{yellow!50}57
& \cellcolor{yellow!50}73.56
& \cellcolor{yellow!50}18.22 \\

& Mutual Information & 1,2,3,4,5,6,7,8,9
& 57 & 69.01 & 22.77 \\

\bottomrule
\end{tabular}
}
\end{minipage}
\hfill
\begin{minipage}[t]{0.48\textwidth}
\centering
\caption{Pruning strategies when removing 11 blocks (91.70\%).}
\label{tab:appendix_remove11}
\resizebox{\columnwidth}{!}{
\begin{tabular}{clcccc}
\toprule
Pruning Ratio
& \multicolumn{1}{c}{Method}
& Removed Block
& GFLOPs
& WAR (\%)
& $\Delta$ (\%) \\
\midrule
\multirow{10}{*}{91.70\%}

& Random & --
& 30 & 62.05 & 29.73 \\

& Mean & 1,2,3,4,5,6,7,8,9,10,11
& 30 & 59.01 & 32.77 \\

& Variance & 2,3,4,5,6,7,8,9,10,11,12
& 30 & 68.81 & 22.97 \\

& L1 Norm & 1,2,3,4,5,6,7,8,9,10,11
& 30 & 59.01 & 32.77 \\

& \cellcolor{yellow!50}Sensitivity
& \cellcolor{yellow!50}2,3,4,5,6,7,8,9,10,11,12
& \cellcolor{yellow!50}30
& \cellcolor{yellow!50}68.81
& \cellcolor{yellow!50}22.97 \\

& L2 Norm & 1,2,3,4,5,6,7,8,9,10,11
& 30 & 59.01 & 32.77 \\

& Standard Deviation & 2,3,4,5,6,7,8,9,10,11,12
& 30 & 68.81 & 22.97 \\

& Kurtosis & 1,2,3,4,5,6,7,8,9,10,12
& 30 & 58.32 & 33.46 \\

\cmidrule(lr){2-6}
& \cellcolor{yellow!50}Information Entropy
& \cellcolor{yellow!50}2,3,4,5,6,7,8,9,10,11,12
& \cellcolor{yellow!50}30
& \cellcolor{yellow!50}68.81
& \cellcolor{yellow!50}22.97 \\

& Mutual Information & 1,2,3,4,5,6,7,8,9,10,11
& 30 & 59.01 & 32.77 \\

\bottomrule
\end{tabular}
}
\end{minipage}

\end{table}

We provide the complete multi-block pruning results under different pruning ratios in Table~\ref{tab:appendix_remove1}-\ref{tab:appendix_remove11}. Highlighted rows indicate the best-performing pruning criteria.


\end{document}